%% file: acl_latex.tex
\pdfoutput=1

\documentclass[11pt]{article}

\usepackage{emnlp2022}
\usepackage{graphicx}
\usepackage{times}
\usepackage{latexsym}
\usepackage{amsmath}
\usepackage{tabularx}
\usepackage{algorithm}
\usepackage{algpseudocode} 
\usepackage{xspace}
\usepackage{booktabs}
\usepackage{multirow}

\usepackage[T1]{fontenc}

\usepackage[utf8]{inputenc}

\usepackage{microtype}

\usepackage{inconsolata}

\newcommand{\baby}{InforMask\xspace}

\title{InforMask: Unsupervised Informative Masking for \\ Language Model Pretraining}

\author{Nafis Sadeq\thanks{\ \ Equal contribution.}\ , Canwen Xu$^{*}$, Julian McAuley \\
   University of California, San Diego \\
    \texttt{\{nsadeq,cxu,jmcauley\}@ucsd.edu}}

\begin{document}
\maketitle
\begin{abstract}
Masked language modeling is widely used for pretraining large language models for natural language understanding (NLU). However, random masking is suboptimal, allocating an equal masking rate for all tokens. In this paper, we propose InforMask, a new unsupervised masking strategy for training masked language models. InforMask exploits Pointwise Mutual Information (PMI) to select the most informative tokens to mask. We further propose two optimizations for InforMask to improve its efficiency. With a one-off preprocessing step, InforMask outperforms random masking and previously proposed masking strategies on the factual recall benchmark LAMA and the question answering benchmark SQuAD v1 and v2.\footnote{The code and model checkpoints are available at \url{https://github.com/NafisSadeq/InforMask}.}

\end{abstract}

\input{Sources/introduction}

\input{Sources/related_work}

\input{Sources/methodology}

\input{Sources/experiments}
\input{Sources/conclusion}

\section*{Limitations}
We conduct experiments to compare \baby to several prior works on better masking strategies by training them for 3 epochs. We also compare a fully trained InformBERT-base model to BERT and RoBERTa. However, one limitation of our paper is due to our limited computational budget, we are not able to scale the experiments for larger model size, larger corpus, or compare all baselines under the full pretraining setting. Also, our InformBERT model is arguably suboptimal, with a relatively small batch size and no hyperparameter tuning or search at all. 

\section*{Ethics Statement}
Similar to BERT or RoBERTa, our model may contain social biases that preexist in the training corpus. Thus, we do not anticipate any major ethical concerns in addition to those identified in language models~\citep{bender2021on}. However, to the best of our knowledge, there is no research on the impact of masking strategies on social biases, which could be an interesting and important direction for future research.

\section*{Acknowledgements}
We would like to thank the anonymous reviewers for their insightful comments. This project is partly supported by NSF Award \#1750063.

\bibliography{anthology,custom}
\bibliographystyle{acl_natbib}

\input{Sources/appendix}

\end{document}

%% file: Sources/introduction.tex
\section{Introduction}

 \begin{figure*}
    \centering
	\includegraphics[width=0.9\linewidth]{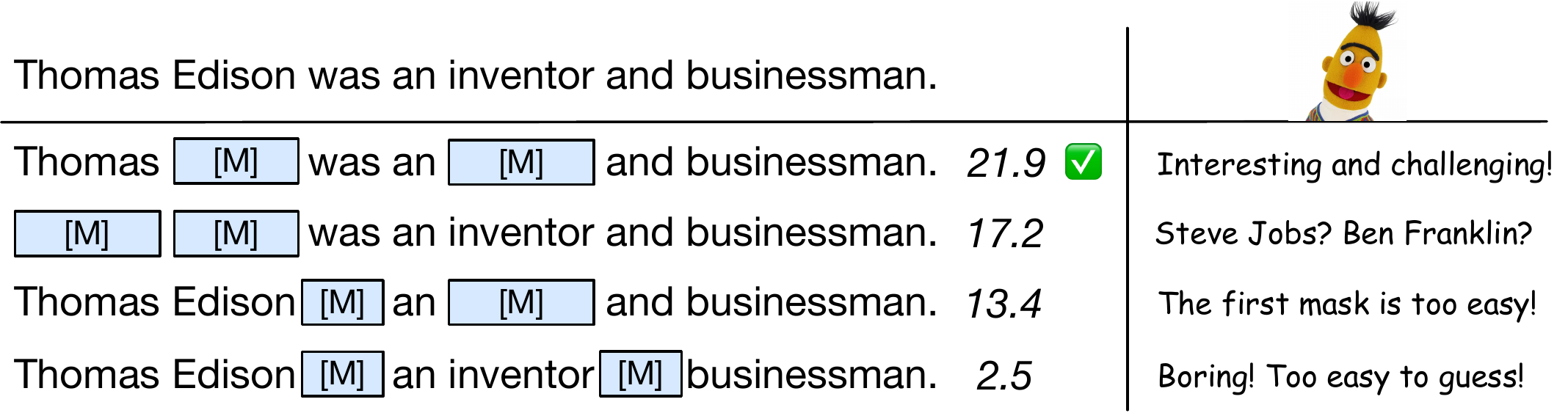}
	\caption{The informative scores of randomly sampled masking candidates ($s=4$). \texttt{[M]} denotes the masked tokens. The pretraining objective of the masked language model (MLM) is to predict the masked tokens based on the context.}
	\label{fig:example}
\end{figure*}

Masked Language Modeling (MLM) is widely used for training language models \cite{devlin2018bert,liu2019roberta,lewis2019bart,2020t5}. MLM randomly selects a portion of tokens from a text sample and replaces them with a special mask token (e.g., \texttt{[MASK]}). However, random masking has a few drawbacks --- it sometimes produces masks that are too easy to guess, providing a small loss that is inefficient for training; some randomly masked tokens can be guessed with only local cues~\citep{joshi2020spanbert}; all tokens have an identical probability to be masked, while (e.g.)~named entities are more important and need special attention~\citep{sun2019ernie,levine2020pmi}.

In this paper, we propose a new strategy for choosing tokens to mask in text samples. We aim to select words with the most information that can benefit the language model, especially for knowledge-intense tasks. To tackle this challenge, we propose \textit{\baby}, an unsupervised informative masking strategy for language model pretraining. First, we introduce \textit{Informative Relevance}, a metric based on Pointwise Mutual Information (PMI, \citealp{fano1961transmission}) to measure the quality of a masking choice. Optimizing this measure ensures the informativeness of the masked token while maintaining a moderate difficulty for the model to predict the masked tokens. This metric is based on the statistical analysis of the corpus, which does not require any supervision or external resource.

However, maximizing the total Informative Relevance of a text sample with multiple masks can be computationally challenging. Thus, we propose a sample-and-score algorithm to reduce the time complexity of masking and diversify the patterns in the output. An example is shown in Figure~\ref{fig:example}. For training a language model with more epochs, we can further accelerate the masking process by only running the algorithm once as a preprocessing step and assigning a token-specific masking rate for each token according to their masking frequency in the corpus, to approximate the masking decisions of the sample-and-score algorithm. After this one-off preprocessing step, masking can be as fast as the original random masking without any further overhead, which can be desirable for large-scale distributed language model training of many epochs.

To verify the effectiveness of our proposed method, we conduct extensive experiments on two knowledge-intense tasks --- factual recall and question answering. On the factual recall benchmark LAMA~\citep{petroni2019language}, \baby outperforms other masking strategies by a large margin. Also, our base-size model, InformBERT, trained with the same corpus and epochs as BERT~\citep{devlin2018bert} outperforms BERT-base on question answering benchmark 
SQuAD~\citep{rajpurkar-etal-2016-squad,rajpurkar-etal-2018-know}. Notably, on the LAMA benchmark, InformBERT outperforms BERT and RoBERTa~\citep{liu2019roberta} models that have 3$\times$ parameters and 10$\times$ corpus size.

To summarize, our contributions are as follows:
\begin{itemize}
    \item We propose \baby, an informative masking strategy for language model pretraining that does not require extra supervision or external resource.
    \item We pretrain and release InformBERT, a base-size English BERT model that substantially outperforms BERT and RoBERTa on the factual recall benchmark LAMA despite having much fewer parameters and less training data. InformBERT also achieves competitive results on the question answering datasets SQuAD v1 and v2.

\end{itemize}

%% file: Sources/related_work.tex
\section{Related Work}

\paragraph{Random Masking}
For pretraining Transformer~\cite{vaswani2017attention} based language models such as BERT~\cite{devlin2018bert}, a portion of the tokens is randomly chosen to be masked to set up the masked language model (MLM) objective. Prior studies have commonly used a masking rate of 15\% \cite{devlin2018bert,joshi2020spanbert,levine2020pmi,sun2019ernie,lan2019albert,he2020deberta}, while some recent studies argue that masking rate of 15\% may be a limitation~\cite{clark2020electra} and the pretraining process may benefit from increasing the masking rate to 40\%~\citep{wettig2022should}. However, random masking is not an ideal choice for learning factual and commonsense knowledge. Words that have high informative value may be masked less frequently compared to (e.g.)~stop words, given their frequencies in the corpus.

\paragraph{Span Masking}
Although random masking is effective for pretraining a language model, some prior works have attempted to optimize the masking procedure.
\citet{joshi2020spanbert} propose SpanBERT where they show improved performance on downstream NLP tasks by masking a span of words instead of individual tokens. They randomly select the starting point of a span, then sample a span size from a geometric distribution and mask the selected span. They continue to mask spans until the target masking rate is met. This paper suggests masking spans instead of single words can prevent the model from predicting masked words by only looking at local cues. However, this masking strategy inevitably reduces the modeling between the words in a span, etc., Mount-Fuji, Mona-Lisa, which may hinder its performance in knowledge-intense tasks.

\paragraph{Entity-based Masking}
Baidu-ERNIE~\citep{sun2019ernie} introduces an informed masking strategy where a span containing named entities will be masked. This approach shows improvement compared to random masking but requires prior knowledge regarding named entities. Similarly, \citet{guu2020retrieval} propose Salient Span Masking where a span corresponding to a unique entity will be masked. They rely on an off-the-shelf named entity recognition (NER) system to identify entity names. LUKE~\citep{yamada2020luke} exploits an annotated entity corpus to explicitly mark out the named entities in the pretraining corpus, and masks non-entity words and named entities separately. 

\paragraph{PMI Masking} \citet{levine2020pmi} propose a masking strategy based on Pointwise Mutual Information (PMI, \citealp{fano1961transmission}), where a span of up to five words can be masked based on the joint PMI of the span of words. PMI-Masking is an adaption of SpanBERT~\citep{joshi2020spanbert} where meaningful spans are masked instead of random ones. However, PMI-Masking only considers correlated spans and fails to focus on unigram named entities. This may lead to suboptimal performance on knowledge intense tasks (details in Section~\ref{subsec:analysis}). In our proposed method, we exploit PMI to determine the informative value of tokens to encourage more efficient training and improve performance on knowledge-intense tasks.

\paragraph{Knowledge-Enhanced LMs}
KnowBERT~\citep{peters2019knowledge} shows that factual recall performance in BERT can be improved significantly by embedding knowledge bases into additional layers of the model. Tsinghua-ERNIE~\citep{zhang2019ernie} exploits a similar approach that injects knowledge graphs into the language model during pretraining. KEPLER~\citep{wang2021kepler} uses a knowledge base to jointly optimizes the knowledge embedding loss and MLM loss on a general corpus, to improve the knowledge capacity of the language model. Similar ideas are also explored in K-BERT~\citep{liu2020k} and CoLAKE~\citep{colake}. CokeBERT~\cite{su2021cokebert} demonstrates that incorporating embeddings for dynamic knowledge context can be more effective than incorporating static knowledge graphs. Other works have attempted to incorporate knowledge in the form of lexical relation~\cite{lauscher2020common}, word sense~\cite{levine2019sensebert}, syntax~\cite{bai2021syntax}, and parts-of-speech (POS) tags~\cite{ke2019sentilare}. However, a high-quality knowledge base is expensive to construct and not available for many languages. Different from these methods, our method is fully unsupervised and does not rely on any external resource.

%% file: Sources/methodology.tex
\section{Methodology}

\baby aims to make masking decisions more `informative'. %
Since not all words are equally rich in information~\citep{levine2020pmi}, we aim to automatically identify more important tokens (e.g., named entities) and increase their probability to be masked while preserving the factual hints to recover them. On the other hand, we would like to reduce the frequency of masking stop words.  Stop words are naturally common in the corpus and they can be important for learning the syntax and structure of a sentence. However, masked stop words can be too easy for a language model to predict, especially in later stages of  LM pretraining. Thus, properly reducing the masking frequency of stop words can improve both the efficiency and performance of the model.

\begin{figure}
\centering
\includegraphics[width=\columnwidth]{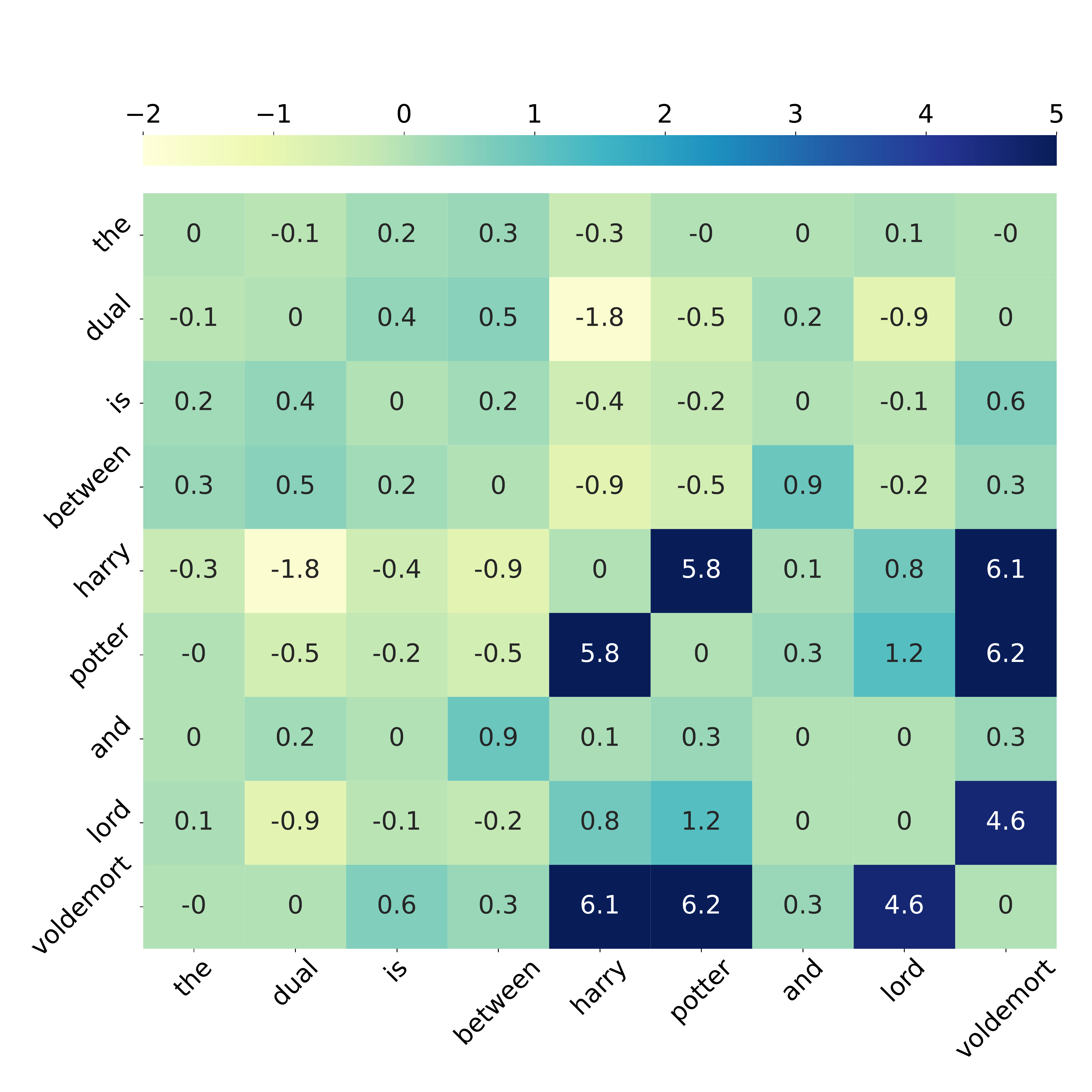}
\caption{The PMI matrix of the words in the sentence `The dual is between Harry Potter and Lord Voldemort.' }
\label{fig:pmi_matrix}
\end{figure}

\begin{algorithm}[t]
	\caption{\baby Algorithm} 
	\label{algo:masking}
	\begin{algorithmic}[1]
	\State $D \gets $ Set of text
	\State $s \gets $ Size of randomly sampled candidates
	\State $F_{i}^{d} \gets $ Informative score for $i$-th masking candidate for text $d$
		\For {$d \in D$}
			\For {$i=1,2,\ldots,s$}
				\State Generate $i$-th masking candidate for $d$
				\State $M_{i}^{d} \gets  $ Masked Tokens
				\State $U_{i}^{d} \gets  $ Unmasked Tokens
				\State $F_{i}^{d} \gets  0 $
				\For{$w_1 \in M_{i}^{d}$}
    				\For{$w_2 \in U_{i}^{d}$}
    				    \State $F_{i}^{d} = F_{i}^{d} + \operatorname{pmi}(w_1,w_2) $ 
    				\EndFor
				
				\EndFor
			\EndFor
			\State Choose candidate with maximum $F_{i}^{d}$
		\EndFor
	\end{algorithmic} 
\end{algorithm}

\subsection{Informative Relevance}
To generate highly informative masking decisions for a sentence, we introduce a new concept, namely Informative Relevance. Informative Relevance is used to measure how relevant a masked word is  to the unmasked words so that it can be meaningful and predictable. The Informative Relevance of a word is calculated by summing up the Pointwise Mutual Information (PMI, \citealp{fano1961transmission}) between the masked word and all unmasked words in the sentence. PMI between two words $w_1$ and $w_2$ represents how `surprising' is the 
co-occurrence between two words,
accounting for their own probabilities. Formally, the PMI of the combination $w_1w_2$ is defined as:
\begin{equation}
    \operatorname{pmi}(w_1,w_2) = \log \frac{p(w_1,w_2)}{p(w_1)p(w_2)}
\end{equation}
The PMI matrix is calculated corpus-wise.
Note that instead of using bigrams (i.e., two words have to be next to each other), we consider the skip-gram co-occurrence within a window. The window size is selected in a way that enables sentence-level co-occurrence to be considered as well as local co-occurrence.

Maximizing the Informative Relevance enables the model to better memorize knowledge and focus on more informative words. Since Informative Relevance is calculated between a masked word and the unmasked words, it also encourages hints to be preserved so that the model can reasonably guess the masked words. As shown in Figure~\ref{fig:pmi_matrix}, the words inside a named entity have a high PMI (e.g., `Harry-Potter' and `Lord-Voldemort') while the two closely related entities also show a high PMI (e.g., Harry-Voldemort). Thus, if we are asked to mask one word, we would mask `Voldemort' since it has the highest Informative Relevance with the remaining words (by summing up the last row or column).

\subsection{Scoring Masking Candidates}
One text sample can have multiple masks. Thus, we define the informative score of a masking decision as the sum of the Informative Relevance of each masked token. However, given the PMI matrix, finding the best $k$ words to mask (i.e., the masking decision with the highest informative score) in a sentence of $n$ words is time-consuming. Iterating all possibilities has 
time complexity  $O(C_n^k)$. By converting it to a minimum cut problem, the time complexity can be reduced to $O(n^2 \log n)$~\citep{stoer1997simple}, which is still 
prohibitive
in practice.

Therefore, we propose to sample $s$ random masking candidates and then rank them by calculating their informative scores. As shown in Figure~\ref{fig:example}, we randomly generate four masking candidates and rank them by their informative scores. We select the candidate with the highest score. This allows us to make a masking decision with time complexity $O(kn)$. Random sampling also introduces more diverse patterns for masking, which could help  training of language models and prevent overfitting. This process is illustrated in Algorithm~\ref{algo:masking}.

\subsection{Token-Specific Masking Rates}
Algorithm~\ref{algo:masking} is already usable by processing the input text on the fly. However, to avoid overfitting, masking should change across epochs. This means we have to run Algorithm 1 every epoch, creating a bottleneck for pretraining. To address this efficiency issue, we use token-specific masking rates to approximate the masking decisions of \baby. Specifically, we generate masks for a corpus using Algorithm 1, and then count the frequency of each token in the vocabulary to be masked as their token-specific masking rates. Note that in this way, Algorithm~\ref{algo:masking} is only executed once, as a prepossessing step. Furthermore, we can use a small portion of the corpus to calculate the token-specific masking rates, making it even faster.\footnote{For the Wikipedia corpus, the average rate of change for token-specific masking rates falls below 0.8\% after processing only 1\% of the corpus.} After this, we can perform random masking, except that every token has its own masking rate.

%% file: Sources/experiments.tex
\begin{table}[t!]
\centering
\begin{tabular}{lrr}
\toprule
Data Subset & \#Relations & \#Samples \\
\midrule
ConceptNet & 1 & 29774 \\
Squad & 1 & 305 \\
GoogleRE & 3 & 4994 \\
TREx & 41 & 34032 \\
\midrule
Total & 46 & 69105\\
\bottomrule
\end{tabular}
\caption{Statistics of LAMA~\citep{petroni2019language}.}
\label{tab:lama_summary}
\end{table}

\begin{table}[t!]
\centering
\resizebox{\columnwidth}{!}{
\begin{tabular}{lrr}
\toprule
Dataset & SQuAD v1 & SQuAD v2 \\
\midrule
\#Examples & 108k & 151k \\
\#Negative Examples & 0 & 54k \\
\#Articles & 536 & 505 \\
\bottomrule
\end{tabular}
}
\caption{Statistics of SQuAD v1 and v2~\citep{rajpurkar-etal-2016-squad,rajpurkar-etal-2018-know}.}
\label{tab:squad_summary}
\end{table}

\begin{table*}
\centering
\resizebox{\textwidth}{!}{
\begin{tabular}{clrrcccccc}
\toprule
& \multirow{2}{*}{Model} & \multirow{2}{*}{\#Param.} & \multirow{2}{*}{Corpus Size} & \multirow{2}{*}{Epochs} & \multicolumn{4}{c}{LAMA~\citep{petroni2019language}} \\
\cmidrule{6-9}
 & & & & & ConceptNet & Squad & GoogleRE & TREx & Overall \\
\midrule
\multirow{4}{*}{(a)} & Random~\shortcite{devlin2018bert} & 125M & 16 GB & 3 & 0.091 & 0.124 & 0.396 & 0.582 & 0.549 \\
& Span~\shortcite{joshi2020spanbert} & 125M & 16 GB & 3 & 0.056 & 0.102 & 0.377 & 0.524 & 0.495 \\
& PMI~\shortcite{levine2020pmi} & 125M & 16 GB & 3 & 0.075 & 0.115 & 0.396 & 0.552 & 0.522 \\
& \baby & 125M & 16 GB & 3 & \textbf{0.109} & \textbf{0.133} & \textbf{0.410} & \textbf{0.627} & \textbf{0.591} \\
\midrule
\multirow{5}{*}{(b)} & BERT-base & 110M & 16 GB & 40 & 0.191 & 0.229 & 0.340 & 0.587 & 0.553 \\
& BERT-large & 340M & 16 GB & 40 & 0.218 & 0.284 & 0.354 & 0.621 & 0.585 \\
& RoBERTa-base & 125M & 160 GB & 40 & 0.223 & 0.307 & 0.423 & 0.630 & 0.592 \\
& RoBERTa-large & 355M & 160 GB & 40 & \textbf{0.260} & 0.329 & 0.435 & 0.672 & 0.632 \\
& InformBERT & 125M & 16 GB & 40 & 0.201 & \textbf{0.384} & \textbf{0.509} & \textbf{0.739} & \textbf{0.698}\\
\bottomrule
\end{tabular}
}
\caption{Performance of different masking strategies and models on LAMA~\citep{petroni2019language}. (a) We compare the models trained with different masking strategies for 3 epochs. (b) We compare InformBERT, a BERT model trained with InforMask for 40 epochs with BERT and RoBERTa models.}
\label{tab:performance_categorized}
\end{table*}

\begin{table*}
\centering
\resizebox{0.84\textwidth}{!}{
\begin{tabular}{clrrccccc}
\toprule
& \multirow{2}{*}{Model} & \multirow{2}{*}{\#Param.} & \multirow{2}{*}{Corpus Size} & \multirow{2}{*}{Epochs} & \multicolumn{2}{c}{SQuAD v1} & \multicolumn{2}{c}{SQuAD v2} \\
\cmidrule(lr){6-7} \cmidrule(lr){8-9}
& &&& & F1 & EM & F1 & EM \\
 \midrule
\multirow{4}{*}{(a)} & Random~\shortcite{devlin2018bert} & 125M & 16 GB & 3  & 79.08 & 69.44 & 66.48 & 63.15 \\
& Span~\shortcite{joshi2020spanbert} & 125M & 16 GB & 3  & 78.88 & 69.04 & 64.95 & 61.38 \\
& PMI~\shortcite{levine2020pmi} & 125M & 16 GB & 3  & 80.31 & 70.98 & 66.25 & 62.82 \\
& \baby & 125M & 16 GB & 3  & \textbf{80.47} & \textbf{71.41} & \textbf{67.29} & \textbf{63.90}\\
\midrule
\multirow{2}{*}{(b)} & BERT-base & 110M & 16 GB & 40  & 81.07 & 88.52 & 72.35 & 75.75 \\
& InformBERT & 125M & 16 GB & 40 & \textbf{81.22} & \textbf{88.61} & \textbf{72.71} & \textbf{75.86} \\
\bottomrule
\end{tabular}
}
\caption{Performance on SQuAD v1 and v2~\citep{rajpurkar-etal-2016-squad,rajpurkar-etal-2018-know} development set.} %
\label{tab:squad}
\end{table*}

\section{Experiments}

\subsection{Experimental Settings}
\paragraph{Pretraining Corpus}
Following BERT~\citep{devlin2018bert}, we use the Wikipedia and Book Corpus datasets available from Hugging Face
\cite{Hugg}. The corpus contains $\sim$3.3B tokens. To be consistent with BERT, we use an overall masking rate of 15\%. The PMI matrix is calculated on the Wikipedia corpus, with a size of 100k $\times$ 100k. Word co-occurrence statistics are computed with a window size of 11. We set the candidate sampling size per document $s$ to 30. It takes $\sim$4 hours to preprocess and generate token-specific masking rates on a 16-core CPU server with 256 GB RAM.

\paragraph{Evaluation Benchmarks}
To evaluate different masking strategies, we use the LAMA benchmark~\cite{petroni2019language} to test the knowledge of the models. LAMA is a probe for analyzing the factual and commonsense knowledge contained in pretrained language models. Thus, it is suitable for evaluating the knowledge learned during pretraining.
LAMA has around 70,000 factual probing samples across 46 factual relations. A summary of the benchmark is shown in Table~\ref{tab:lama_summary}. We use Mean Reciprocal Rank (MRR) as the metric for factual recall performance. 

In addition to the knowledge probing task, we also conduct experiments on real-world question answering datasets, which requires commonsense knowledge as well. We conduct experiments on SQuAD v1 and v2~\citep{rajpurkar-etal-2016-squad,rajpurkar-etal-2018-know} and report the F1 and Exact Match (EM) scores on the development set. The statistics of the benchmark are shown in Table~\ref{tab:squad_summary}. We provide additional results on GLUE~\citep{wang2018glue} benchmark in Appendix~\ref{sec:glue}.  %

\paragraph{Baselines} We compare InforMask in two settings: \textbf{(a)} We use the same tokenizer and hyperparameters to pretrain BERT random masking~\citep{devlin2018bert}, SpanBERT~\citep{joshi2020spanbert} and PMI-Masking~\citep{levine2020pmi} for 3 epochs. The choice of 3 epochs is according to our limited computational budget. \textbf{(b)} We continue training InforMask until 40 epochs. The 40-epoch model is denoted as \textit{InformBERT}. We compare InformBERT to BERT-base~\citep{devlin2018bert}, which is trained with the same corpus for 40 epochs as well. We also include results of BERT-large and RoBERTa for reference, though they are either larger in size or trained with more data and thus are not directly comparable.

\paragraph{Training Details} Our implementation is based on Hugging Face Transformers~\citep{hftransformer}. We train the baselines and our model with 16 Nvidia V100 32GB GPU. For our model and all baselines trained, we use a fixed vocabulary size of 50,265. The model architecture is a base-size BERT model, with 12 Transformer layers with 12 attention heads. The hidden size is set to 768. The overall batch size is 256. We use an AdamW optimizer~\citep{adamw} with a learning rate of 5e-5. Note that we do not perform any hyperparameter searching or tuning for any model (including InformBERT) given our limited computational budget.

\subsection{Experimental Results}

\paragraph{Impact of Masking Strategies} 
We conduct a fair comparison among different masking strategies, using the same tokenization and hyperparameters. As shown in Table~\ref{tab:performance_categorized}(a), \baby outperforms other masking strategies by a large margin on all subsets of LAMA~\citep{petroni2019language}. As shown in Table~\ref{tab:squad}(a), on both SQuAD v1 and v2~\citep{rajpurkar-etal-2016-squad,rajpurkar-etal-2018-know}, \baby outperforms other masking strategies. Notably, PMI-Masking achieves higher performance on SQuAD while underperforming random masking on LAMA (to be detailed shortly) but our \baby achieves better results on both of them.

\begin{figure}[t!]
\centering
\includegraphics[width=\columnwidth]{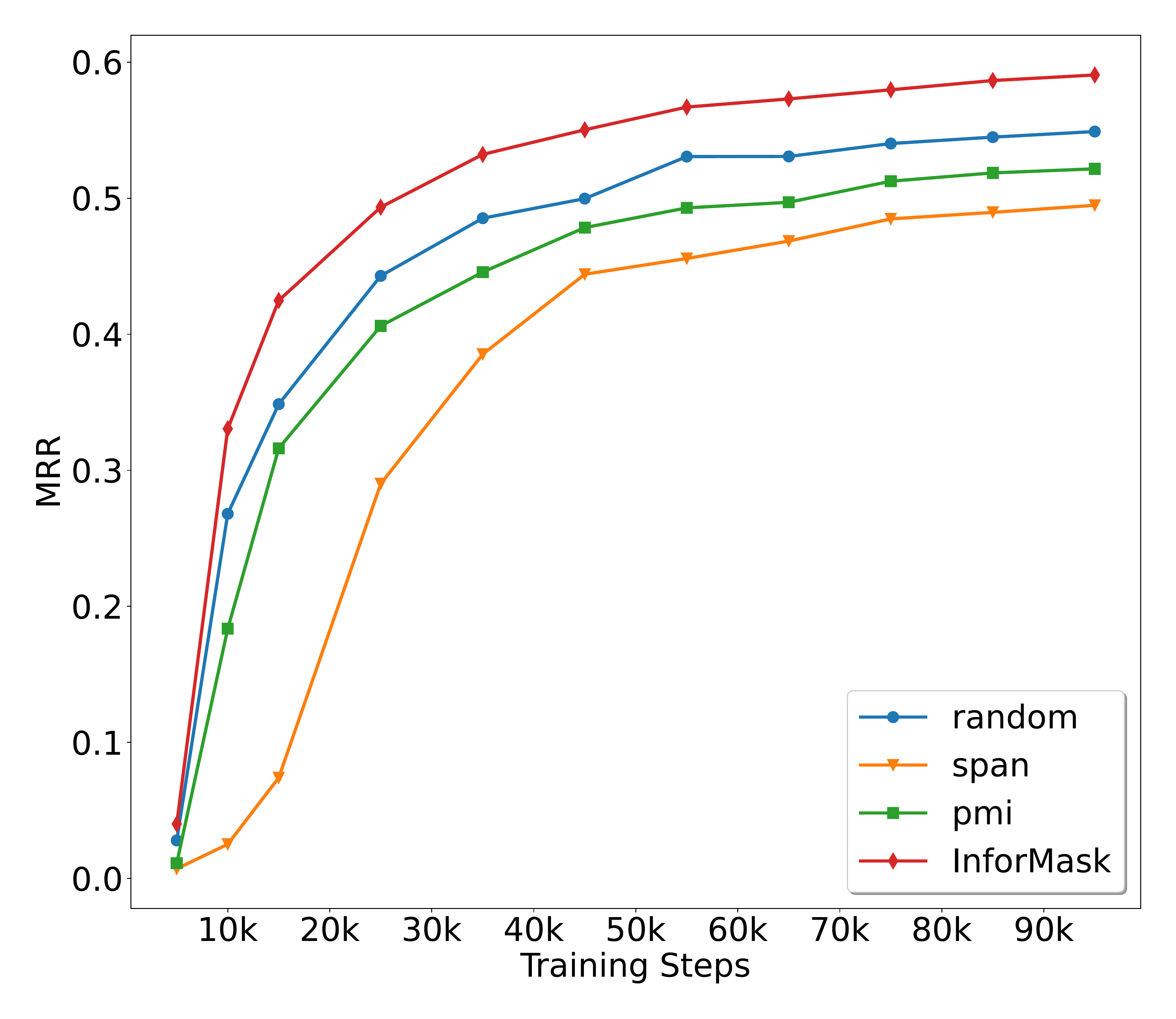}
\caption{Macro average MRR of different masking strategies on LAMA, evaluated every 10k steps.}
\label{fig:masking_policy_graph}
\end{figure}

Also, we compare our 40-epoch InformBERT model with BERT and RoBERTa models. As shown in Table~\ref{tab:performance_categorized}(b), InformBERT outperforms the BERT model trained with the same epochs and corpus by 0.145 overall. It also achieves higher performance than RoBERTa-base, despite being trained with 10\% of RoBERTa's corpus size. To our surprise, it also outperforms both BERT-large and RoBERTa-large, with only 1/3 parameters. The breakdown of performance for each relation can be found in Appendix~\ref{sec:lama_performance_details}. Moreover, InformBERT outperforms BERT-base for fine-tuning on SQuAD v1 and v2, demonstrating its capability for downstream question answering, as shown in Table~\ref{tab:squad}(b).

\begin{figure}[t!]
\centering
\includegraphics[width=\columnwidth]{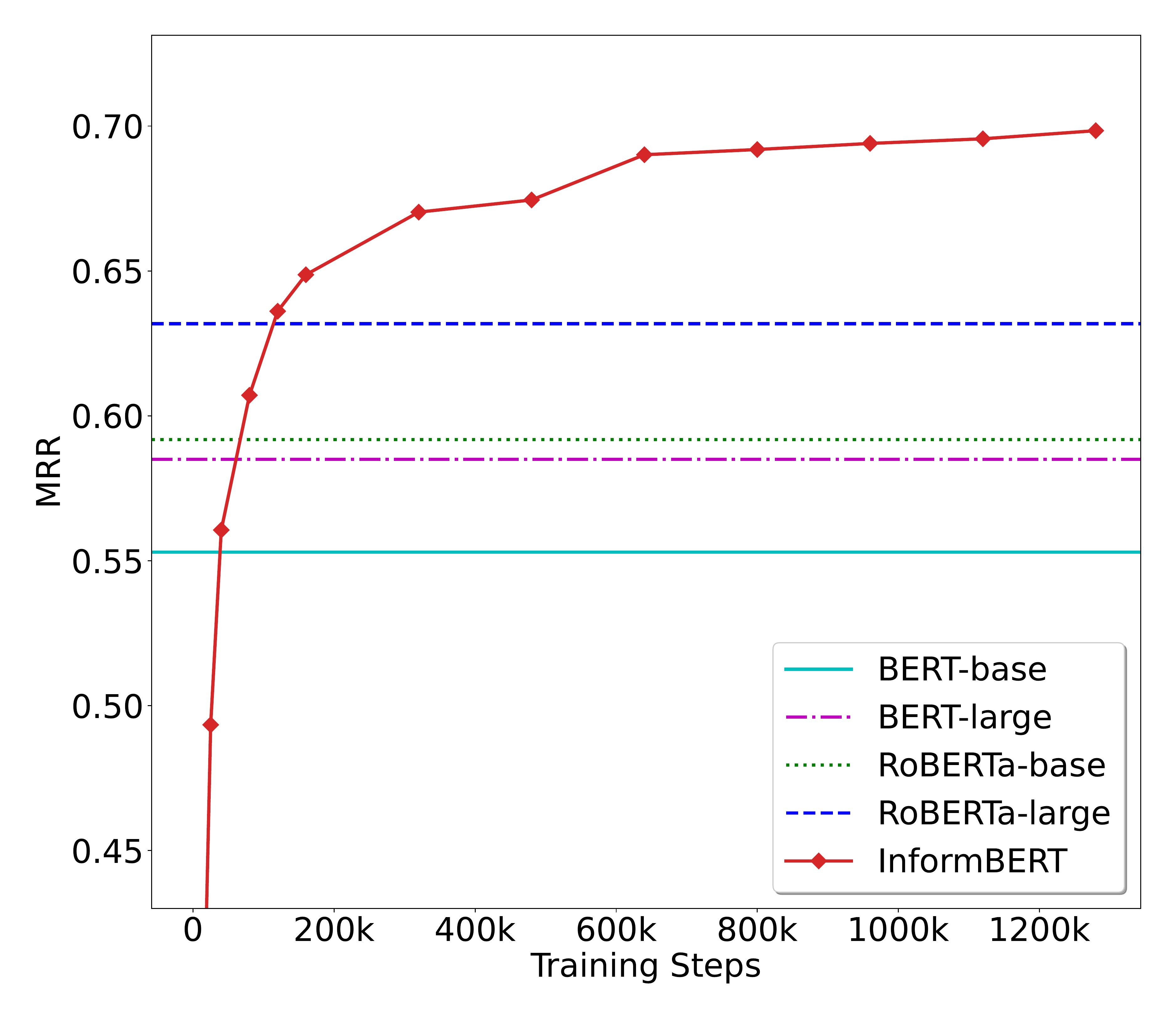}
\caption{Performance of InformBERT for the full pretraining process. It achieves comparable performance with BERT-base after 40k training steps and even RoBERTa-large after 120k training steps.}
\label{fig:step_vs_mrr_full}
\end{figure}

\paragraph{Training Dynamics} As shown in Figure~\ref{fig:masking_policy_graph}, \baby demonstrates an outstanding training efficiency. \baby outperforms other masking strategies from the beginning of the training process and keeps the lead through the training. Notably, span masking and PMI-Masking underperform random masking, indicating their inability on the knowledge-intense task. Span masking also significantly underperforms other masking strategies in the early stage of pretraining, suggesting it may take longer to train the model. For the entire pretraining process, as shown in Figure~\ref{fig:step_vs_mrr_full}, the model trained with \baby outperforms BERT and RoBERTa with fewer than $\sim$15\% of the training steps, verifying the efficiency of our masking strategy.

\begin{figure}[t]
\centering
\includegraphics[width=\columnwidth]{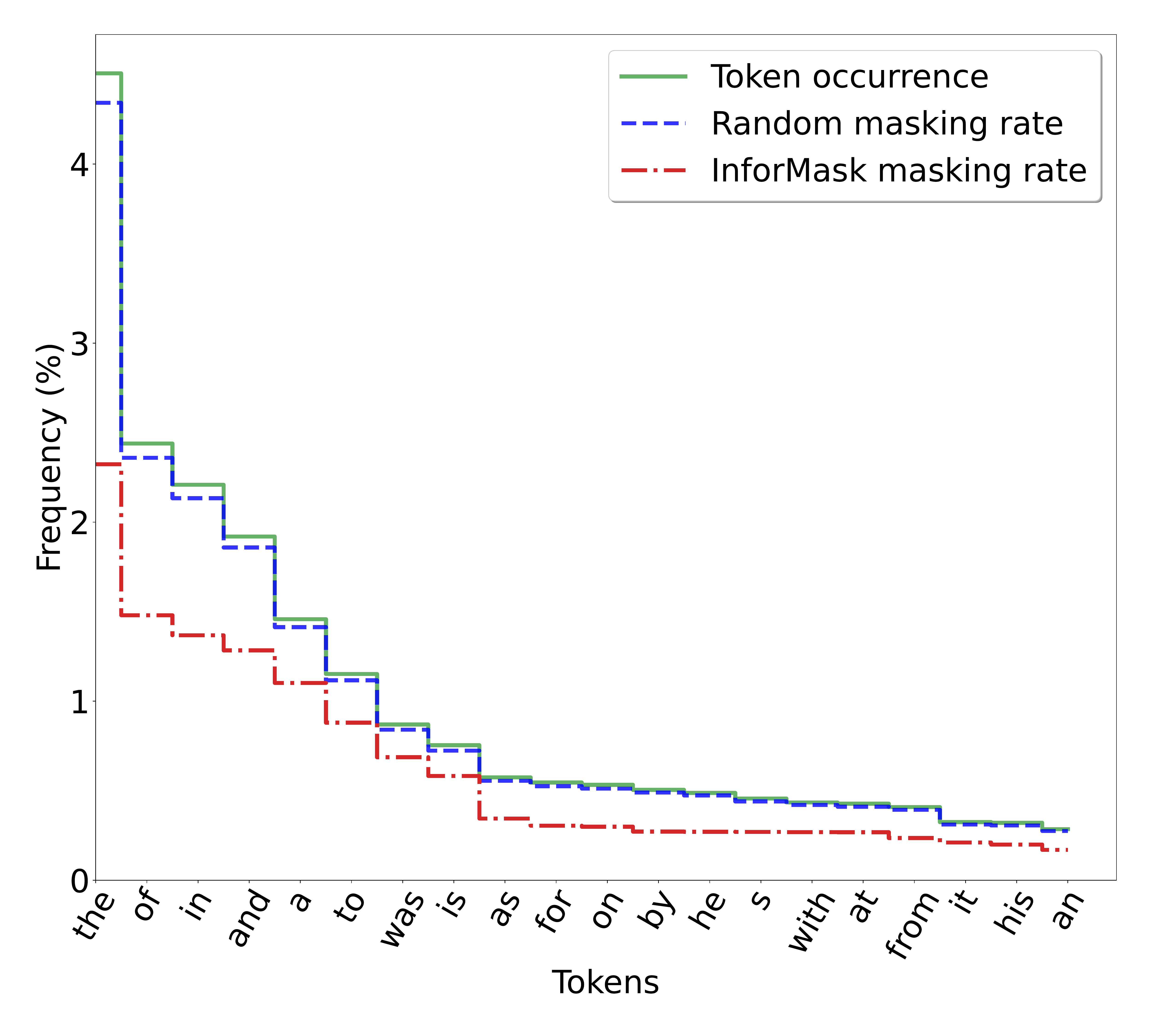}
\caption{Frequency of common stop words and their corresponding masking rates by \baby.}
\label{fig:mask_freq}
\end{figure}

\paragraph{Impact on Stop Words and Entities}
As shown in Figure~\ref{fig:mask_freq}, without explicitly specifying the stop words, \baby can identify the stop words and reduce their probability to be masked. \baby can also automatically increase the masking probability of named entities. The average masking probability of named entities is 0.25 with a standard deviation of 0.07, while the overall masking probability of all tokens is around 0.15.\footnote{We use an off-the-shelf named entity recognition system to verify the effectiveness of our approach only. It is not a necessary component of the proposed system.} This allows the model to focus on more important tokens and maintain an appropriate difficulty of prediction, facilitating the pretraining process.

\begin{figure}[t]
\centering
\includegraphics[width=\columnwidth]{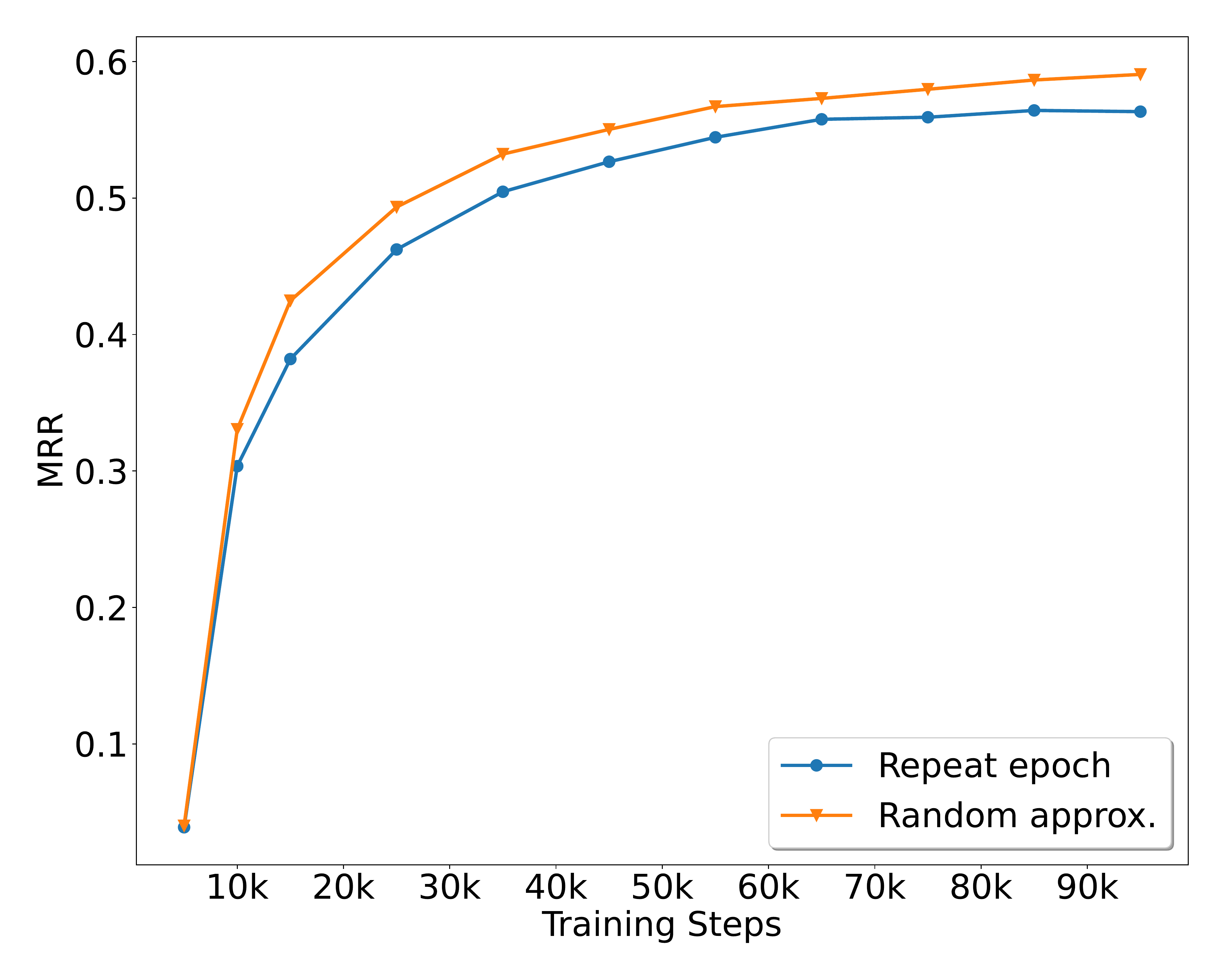}
\caption{Comparison between looping the same data and using token-specific masking rate to approximate the masking decisions. The models are trained for 3 epochs.}
\label{fig:static_vs_dynamic}
\end{figure}

\paragraph{Impact of Token-Specific Masking Rates} As mentioned before, the use of token-specific masking rate can enormously save time and RAM for data processing, as spending hours of processing for each epoch can be infeasible and becomes a bottleneck for distributed training. Another possible solution is to loop the same masked data for every epoch. Thus, we conduct an experiment to compare the two solutions: approximation and repetition. Note that for simplicity, the token-specific masking rate is applied from the first epoch. As shown in Figure~\ref{fig:static_vs_dynamic}, our approximation strategy keeps outperforming the repetition strategy even in the first epoch. As we analyze, this can be attributed to the more diverse patterns introduced during the approximation. Also, the performance of the model trained with the repetition strategy converges or even slightly declines after 60k training steps while the performance of the model trained with approximation keeps increasing.

\begin{figure}
\centering
\includegraphics[width=\columnwidth]{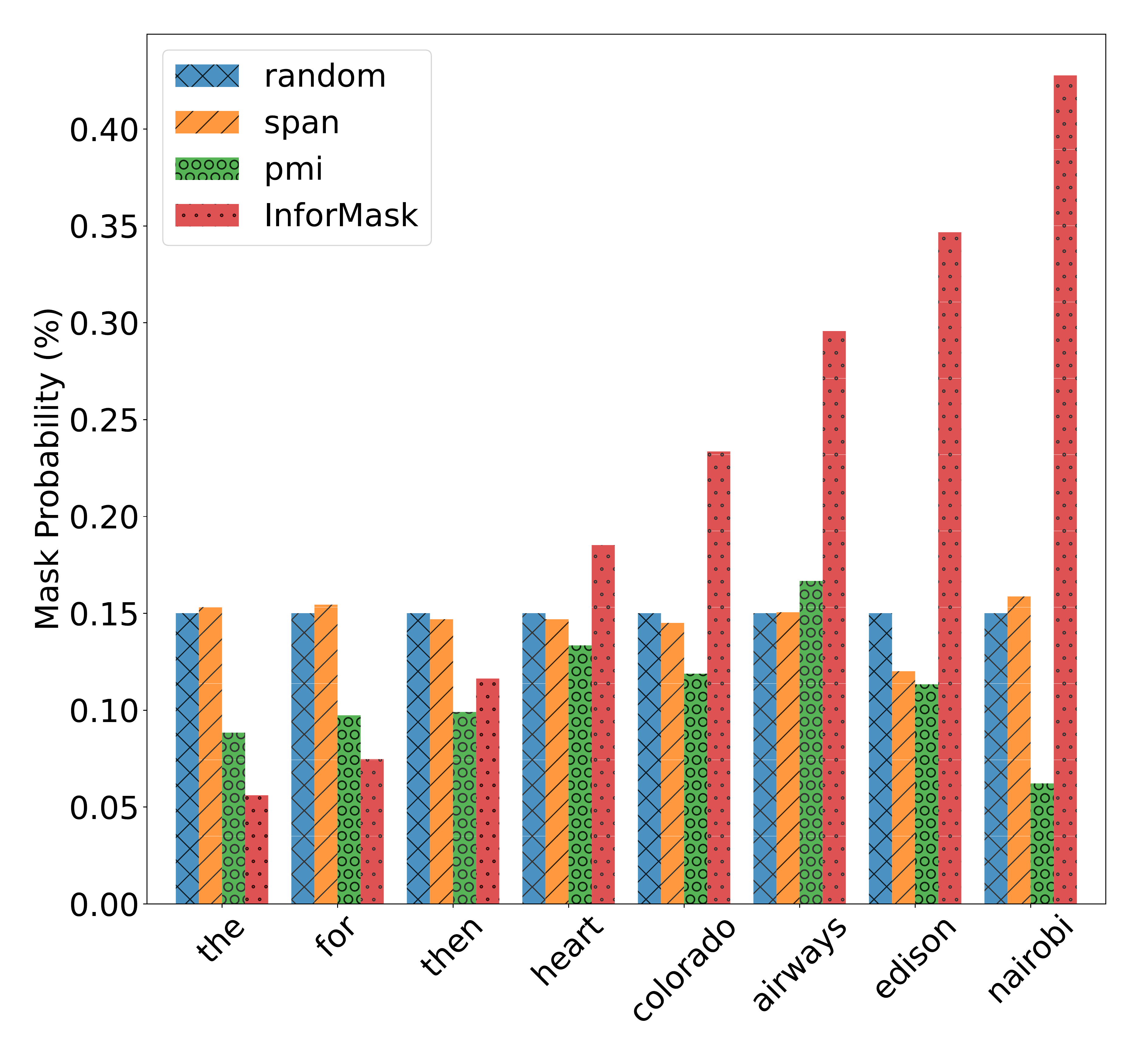}
\caption{Masking rate of tokens according to different masking policies.}
\label{fig:masking_rate_vs_policy}
\end{figure}

\begin{table*}
\resizebox{\textwidth}{!}{
\begin{tabular}{llllll}
\toprule
\multirow{2}{*}{Query} & \multirow{2}{*}{Ground Truth} & \multicolumn{2}{c}{InformBERT} & \multicolumn{2}{c}{RoBERTa-base} \\
\cmidrule(lr){3-4} \cmidrule(lr){5-6}
 &  & Prediction & Score & Prediction & Score \\
\midrule
 &  & france & 0.09 & montreal & 0.12 \\
Antoine Coypel was born in \texttt{[MASK]}. & paris & \textbf{paris} & 0.08 & toronto & 0.03 \\
 &  & haiti & 0.04 & \textbf{paris} & 0.03 \\
 \midrule
 &  & \textbf{espn} & 0.20 & cbs & 0.18 \\
SpeedWeek is an American television program on \texttt{[MASK]}. & espn & nbc & 0.10 & cnbc & 0.13 \\
 &  & mtv & 0.09 & spike & 0.10 \\
 \midrule
 &  & \textbf{microsoft} & 0.20 & intel & 0.06 \\
Phil Harrison is a corporate vice president of \texttt{[MASK]}. & microsoft & ibm & 0.15 & ibm & 0.05 \\
 &  & motorola & 0.05 & \textbf{microsoft} & 0.03 \\ 
 \midrule
 &  & \textbf{french} & 0.43 & young & 0.13 \\
Laurent Casanova was a \texttt{[MASK]} politician. & french & canadian & 0.32 & \textbf{french} & 0.09 \\
 &  & haitian & 0.05 & successful & 0.04\\ 
 \midrule
  &  & \textbf{bishops} & 0.13 & men & 0.17 \\
The chief administrators of the church are \texttt{[MASK]}. & bishops & priests & 0.07 & christians & 0.09 \\
 &  & appointed & 0.06 & women & 0.08\\ 
 \bottomrule
\end{tabular}
}
\caption{Some examples of InformBERT and RoBERTa-base predictions on LAMA~\citep{petroni2019language}. We show the queries and the ground-truth answers with the model predictions. We only show the top-3 predictions made by each model.}
\label{tab:knowledge_probe}
\end{table*}

\label{subsec:analysis}
\paragraph{\baby vs. PMI Masking} PMI Masking~\cite{levine2020pmi} uses PMI to mask a span of correlated tokens. A named entity often constitutes a correlated span and therefore, is more likely to be masked in PMI-Masking. As mentioned before, we observe that PMI-Masking performs worse than random masking on LAMA (see Figure~\ref{fig:masking_policy_graph}).

To investigate the reason, we compute the individual masking rates of some tokens according to each masking policy. 
As shown in Figure~\ref{fig:masking_rate_vs_policy}, we can see that PMI-Masking increases the masking rate of tokens that are part of correlated spans. However, it decreases the masking rate of tokens that are not within any correlated span, even if that token is a named entity. Consider the token `Airways' for example. This token may be part of a correlated span such as `British Airways' or `Qatar Airways'. PMI-Masking, therefore, increases the masking rate of this token compared to random masking. On the other hand, the tokens `Colorado' and `Nairobi', which are unigram named entities, are less likely to be masked, compared to random masking. Given that the overall masking rate is fixed and PMI-Masking favors correlated spans, the masking rates of `Colorado' and `Nairobi' inevitably get lower. This can be the reason behind PMI-Masking's failure.

In contrast, \baby uses PMI to compute the individual Informative Relevance of tokens. It can increase the masking rate of tokens with high informative saliency, regardless of whether they are part of a correlated span or not. This helps \baby achieve superior factual recall performance.

\subsection{Case Study}

Table~\ref{tab:knowledge_probe} shows the example knowledge probes and answers produced by InformBERT and RoBERTa. For the query `SpeedWeek is an American television program on \texttt{[MASK]}.', RoBERTa is unable to produce the correct answer in the top-3 predictions. But InformBERT correctly predicts `ESPN' to be the top candidate. Similarly, InformBERT correctly predicts the answer `bishops' for the query `The chief administrators of the church are \texttt{[MASK]}.' RoBERTa is unable to predict the answer and produces more generic words such as `men', `women', and `Christians'.

We summarize the errors into two notable categories. They are relevant for all the models involved, not just InformBERT. First, we observe that many errors involve rare named entities. Some named entities are less frequent so the model is unable to learn anything useful about them, or they occur so rarely that they do not even appear in the language model vocabulary. We found that around 19\% of the errors made by our model on the LAMA benchmark is associated with out-of-vocabulary tokens. Second, it is challenging for a language model to predict the granularity of the fact being asked or distinguish it from an alternate fact that may hold for a query. For the example query `Antoine Coypel was born in \texttt{[MASK]}.', the LAMA dataset has only one true label `Paris'. In this example, InformBERT prefers the name of the country (`France') over the name of a city (`Paris'). This confusion is related to the granularity of location and both answers can be considered correct. However, it is being classified as an error because the labels in the test set are not comprehensive.

Another type of confusion can be found for RoBERTa with the query `Laurent Casanova was a \texttt{[MASK]} politician.'. The model is trying to decide whether to use the adjective `young', `French', or `successful'. In theory, these three adjectives may be valid simultaneously for the same entity. It can be challenging for the language model to pick the expected one in the context. We include more examples of knowledge probes with InformBERT in Appendix~\ref{sec:lama_more_examples}.

%% file: Sources/conclusion.tex
\section{Conclusion}
In this work, we propose \baby, an unsupervised masking policy that masks tokens based on their informativeness. \baby achieves superior performance in knowledge-intense tasks including factual recall and question answering. We explore the impact of different masking strategies on learning factual and commonsense knowledge from pretraining and analyze why previously proposed masking techniques are suboptimal. For future work, we would like to scale up the pretraining and explore more factors for knowledge acquisition during unsupervised text pretraining.

%% file: Sources/appendix.tex
\newpage
\appendix
\onecolumn
\section{Performance Breakdown on LAMA}
\label{sec:lama_performance_details}
\begin{table*}[h!]
\centering
\resizebox{0.9\textwidth}{!}{
\begin{tabular}{lllllll}
\toprule
Subset & Relation & BERT-base & BERT-large & RoBERTa-base & RoBERTa-large & InformBERT \\
\midrule
ConceptNet & test & 0.191 & 0.218 & 0.223 & \textbf{0.260} & 0.201 \\
GoogleRE & dateOfBirth & 0.108 & 0.115 & 0.092 & 0.108 & \textbf{0.122} \\
GoogleRE & placeOfBirth & 0.475 & 0.493 & 0.610 & 0.612 & \textbf{0.732} \\
GoogleRE & placeOfDeath & 0.388 & 0.403 & 0.528 & 0.582 & \textbf{0.607} \\
Squad & test & 0.229 & 0.284 & 0.307 & 0.329 & \textbf{0.384} \\
TREx & P1001 & 0.786 & 0.817 & 0.810 & 0.846 & \textbf{0.881} \\
TREx & P101 & 0.453 & 0.499 & 0.307 & 0.380 & \textbf{0.507} \\
TREx & P103 & 0.842 & 0.876 & 0.841 & 0.857 & \textbf{0.907} \\
TREx & P106 & 0.656 & 0.675 & 0.540 & 0.599 & \textbf{0.674} \\
TREx & P108 & 0.584 & 0.596 & 0.658 & \textbf{0.725} & 0.704 \\
TREx & P127 & 0.546 & 0.570 & 0.661 & 0.688 & \textbf{0.743} \\
TREx & P1303 & 0.387 & 0.442 & 0.233 & 0.277 & \textbf{0.445} \\
TREx & P131 & 0.650 & 0.685 & 0.742 & 0.778 & \textbf{0.867} \\
TREx & P136 & 0.621 & 0.666 & 0.557 & 0.596 & \textbf{0.675} \\
TREx & P1376 & 0.730 & 0.768 & 0.631 & 0.630 & \textbf{0.840} \\
TREx & P138 & 0.509 & 0.533 & 0.515 & 0.548 & \textbf{0.742} \\
TREx & P140 & 0.606 & 0.674 & 0.668 & 0.728 & \textbf{0.751} \\
TREx & P1412 & 0.777 & 0.801 & 0.799 & 0.824 & \textbf{0.860} \\
TREx & P159 & 0.468 & 0.486 & 0.660 & 0.701 & \textbf{0.789} \\
TREx & P170 & 0.860 & 0.886 & 0.878 & 0.908 & \textbf{0.928} \\
TREx & P176 & 0.687 & 0.731 & 0.717 & 0.770 & \textbf{0.777} \\
TREx & P178 & 0.631 & 0.683 & 0.711 & \textbf{0.744} & 0.721 \\
TREx & P19 & 0.424 & 0.441 & 0.620 & 0.652 & \textbf{0.760} \\
TREx & P190 & 0.267 & 0.312 & 0.486 & 0.542 & \textbf{0.662} \\
TREx & P20 & 0.516 & 0.553 & 0.675 & 0.703 & \textbf{0.791} \\
TREx & P264 & 0.273 & 0.300 & 0.003 & 0.005 & \textbf{0.380} \\
TREx & P27 & 0.767 & 0.796 & 0.853 & 0.884 & \textbf{0.895} \\
TREx & P276 & 0.549 & 0.577 & 0.646 & 0.682 & \textbf{0.824} \\
TREx & P279 & 0.554 & 0.589 & 0.512 & 0.560 & \textbf{0.594} \\
TREx & P30 & 0.832 & 0.868 & 0.845 & 0.896 & \textbf{0.918} \\
TREx & P31 & 0.650 & 0.665 & 0.597 & 0.631 & \textbf{0.652} \\
TREx & P36 & 0.425 & 0.447 & 0.484 & 0.511 & \textbf{0.758} \\
TREx & P361 & 0.554 & 0.596 & 0.442 & 0.480 & \textbf{0.607} \\
TREx & P364 & 0.738 & 0.767 & 0.661 & 0.704 & \textbf{0.811} \\
TREx & P37 & 0.734 & 0.766 & 0.711 & 0.743 & \textbf{0.788} \\
TREx & P39 & 0.615 & 0.647 & 0.501 & 0.550 & \textbf{0.636} \\
TREx & P407 & 0.648 & 0.705 & 0.665 & \textbf{0.710} & 0.695 \\
TREx & P413 & 0.480 & 0.501 & 0.508 & \textbf{0.564} & 0.508 \\
TREx & P449 & 0.470 & 0.473 & 0.652 & 0.685 & \textbf{0.735} \\
TREx & P463 & 0.676 & 0.692 & 0.641 & 0.683 & \textbf{0.736} \\
TREx & P47 & 0.532 & 0.582 & 0.606 & 0.628 & \textbf{0.860} \\
TREx & P495 & 0.707 & 0.737 & 0.805 & \textbf{0.855} & 0.823 \\
TREx & P527 & 0.499 & 0.571 & 0.492 & \textbf{0.585} & 0.575 \\
TREx & P530 & 0.448 & 0.493 & 0.740 & \textbf{0.812} & 0.802 \\
TREx & P740 & 0.343 & 0.369 & 0.672 & 0.715 & \textbf{0.731} \\
TREx & P937 & 0.554 & 0.587 & 0.720 & 0.741 & \textbf{0.797} \\
\bottomrule
\end{tabular}
}
\caption{Relation by relation performance comparison on LAMA~\citep{petroni2019language}.}
\label{tab:relation_by_relation}
\end{table*}

\newpage

\section{More LAMA Examples}
\label{sec:lama_more_examples}

\begin{table*}[h]
    \resizebox{\textwidth}{!}{
    \begin{tabular}{lll}
    \toprule
    Query & Ground truth & Top predictions (with confidence) \\
    \midrule
    Communicating is for gaining \texttt{[M]}. & knowledge & knowledge(0.22), information(0.09), insight(0.04) \\
    Competing against someone requires a desire to \texttt{[M]}. & win & compete(0.35), win(0.25), fight(0.08) \\
    Going on the stage is for performing an \texttt{[M]}. & act & act(0.65), opera(0.2), improvisation(0.02) \\
    Playing is a way to \texttt{[M]} social skills. & learn & learn(0.22), develop(0.15), improve(0.14) \\
    Gallagher was born on 14 December 1978 in \texttt{[M]} . & scotland & ireland(0.25), scotland(0.07), dublin(0.05) \\
    Crisp died at her home in \texttt{[M]}, Arizona . & phoenix & tucson(0.34), phoenix(0.12), prescott(0.12) \\
    Frank Marion died in 1963 in \texttt{[M]}, Connecticut . & stamford & hartford(0.12), stamford(0.1), middletown(0.1) \\
    Mattingly died in 1951 in \texttt{[M]}, Kentucky . & louisville & louisville(0.4), lexington(0.14), ashland(0.03) \\
    Smith died on 26 February 1832 in \texttt{[M]} . & london & england(0.08), london(0.07), ireland(0.03) \\
    Newton played as \texttt{[M]} during Super Bowl 50. & quarterback & quarterback(0.09), referee(0.05), mvp(0.05) \\
    Warsaw is the most diverse \texttt{[M]} in Poland. & city & city(0.63), town(0.13), settlement(0.03) \\
    Quran is a \texttt{[M]} text. & religious & religious(0.21), muslim(0.1), biblical(0.08) \\
    president, and Thomas Watson, founder of \texttt{[M]}. & ibm & ibm(0.21), microsoft(0.02), motorola(0.02) \\
    Letham is a village in \texttt{[M]}, Scotland. & angus & fife(0.52), angus(0.24), highland(0.06) \\
    Hugh Ragin is an American \texttt{[M]} trumpeter. & jazz & jazz(0.97), classical(0.01), rock(0.01) \\
    Avishkaar is a 1974 \texttt{[M]} movie. & hindi & bollywood(0.31), hindi(0.29), malayalam(0.11) \\
    West of Bern, the population generally speaks \texttt{[M]}. & french & german(0.72), french(0.13), italian(0.05) \\
    He was succeeded as \texttt{[M]} by Christoph Ahlhaus. & mayor & chancellor(0.1), bishop(0.09), mayor(0.05) \\
    His son Hugh became \texttt{[M]} of Saint-Gilles. & abbot & bishop(0.48), abbot(0.28), archbishop(0.13) \\
    During his terms Romania joined \texttt{[M]}. & nato & nato(0.25), yugoslavia(0.15), czechoslovakia(0.07) \\
    It seized \texttt{[M]} and Czechoslovakia in 1938 and 1939. & austria & hungary(0.3), poland(0.23), austria(0.11) \\
    Hostage Life was a Canadian punk band from \texttt{[M]}. & toronto & toronto(0.25), vancouver(0.12), montreal(0.11) \\
    \bottomrule
    \end{tabular}
    }
    \caption{More factual probe examples of InformBERT on LAMA~\citep{petroni2019language}. \texttt{[M]} denotes the masked token.}
    \label{tab:factual_probe}
    \end{table*}

\section{GLUE Performance}
We have conducted additional experiments on GLUE~\citep{wang2018glue}. InformBERT outperforms BERT-base on six out of nine tasks. Notably, InformBERT seems to underperform BERT by a large margin on CoLA, which is focused on the grammatical correctness. We suspect this is because InformBERT pays less attention to stop words that can be important for this task.
\label{sec:glue}
\begin{table*}[h]
\centering
\begin{tabular}{lcccccccccc}
\toprule
 \multirow{2}{*}{Model} & \multicolumn{9}{c}{GLUE~\cite{wang2018glue}} \\
\cmidrule{2-10}
 & CoLA & SST-2 & MRPC & STS-B & QQP & MNLI & QNLI & RTE & WNLI \\
\midrule
BERT-base & \textbf{56.53} & 92.32 & 84.07 & 88.64 & 90.71 & \textbf{83.91} & \textbf{90.66} & 65.57 & 56.34 \\
InformBERT & 52.16 & \textbf{92.66} & \textbf{87.50} & \textbf{88.75} & \textbf{90.90} & 83.13 & 89.82 & \textbf{65.70} & \textbf{56.93} \\ 
\bottomrule
\end{tabular}
\caption{Comparison of InformBERT and BERT-base on the dev. set of GLUE~\citep{wang2018glue}. both models are trained for 40 epochs using the same corpus. We report Matthews correlation for CoLA, Pearson correlation for STS-B and accuracy for other tasks.}
\label{tab:glue}
\end{table*}